\documentclass{article}

\usepackage{microtype}
\usepackage{graphicx}
\usepackage{subfigure}
\usepackage{booktabs} 
\usepackage{makecell}

\usepackage{hyperref}

\usepackage[accepted]{icml2019}

\icmltitlerunning{Oktoberfest Food Dataset}

\begin{document}

\twocolumn[
\icmltitle{Oktoberfest Food Dataset}



\icmlsetsymbol{equal}{*}

\begin{icmlauthorlist}
\icmlauthor{Alexander Ziller}{equal,tum}
\icmlauthor{Julius Hansjakob}{equal,tum}
\icmlauthor{Vitalii Rusinov}{equal,tum}
\icmlauthor{Daniel Z\"ugner}{tum}
\icmlauthor{Peter Vogel}{ilass}
\icmlauthor{Stephan G\"unnemann}{tum}
\end{icmlauthorlist}

\icmlaffiliation{tum}{Department of Computer Science, Technical University of Munich, Munich, Germany}
\icmlaffiliation{ilass}{Ilass AG, Munich, Germany}

\icmlcorrespondingauthor{Alexander Ziller}{alex.ziller@tum.de}
\icmlcorrespondingauthor{Julius Hansjakob}{jhansjakob@googlemail.com}
\icmlcorrespondingauthor{Vitalii Rusinov}{vitaliy.rusinov.cv@gmail.com}

\icmlkeywords{Machine Learning, ICML, Deep Learning, Object Detection, Food}

\vskip 0.3in
]



\printAffiliationsAndNotice{\icmlEqualContribution} 

\begin{abstract}
We release a realistic, diverse, and challenging dataset for object detection on images. The data was recorded at a beer tent in Germany and consists of 15 different categories of food and drink items. We created more than 2,500 object annotations by hand for 1,110 images captured by a video camera above the checkout. We further make available the remaining 600GB of (unlabeled) data containing days of footage. Additionally, we provide our trained models as a benchmark.
Possible applications include automated checkout systems which could significantly speed up the process.
\end{abstract}

\section{Introduction}
In the context of automated checkout systems, it is necessary to know what objects are on the counter and how many of each. Dedicated approaches to count items of different classes using deep learning are still in development. 
Regression-based approaches e.g. \cite{onoro2016towards} do not handle different object classes.

Another group of algorithms is based on object detection, which also yields the location of the objects and makes the visualization of results easier. Object detection can be used naturally to count objects by simply counting the number of high-confidence object detections of interest.

\begin{figure}[ht]
    \centering
    \includegraphics[width=0.45\textwidth]{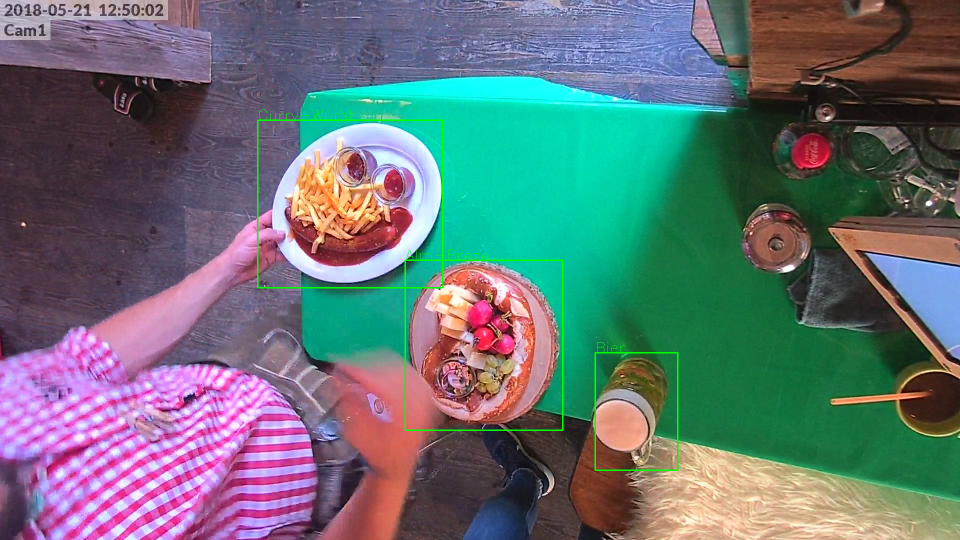}
    \caption{Example image from our dataset. From left to right: Curry Wurst, Alm Breze, Bier.}
    \label{fig:example}
\end{figure}

\section{Dataset}
The source video data, 600 GB, displays the counter at a beer tent during eleven days from four different angles. 
We extracted and annotated images from one of these four cameras.

\subsection{(Annotated) Data Statistics}
{\color{red} }

\begin{tabular}{l|l|l|l}
    Class & Images & Annotations & 
    Average quantity\\ \hline
    Bier & 300 & 436 & 1.45 \\
    Bier Mass & 200 & 299 & 1.50 \\
    Weissbier & 229 & 298 & 1.30 \\
    Cola & 165 & 210 & 1.27 \\
    Wasser & 198 & 284 & 1.43 \\
    Curry-Wurst & 120 & 159 & 1.32 \\
    Weisswein & 81 & 105 & 1.30 \\
    A-Schorle & 90 & 98 & 1.09 \\
    Jaegermeister & 43 & 152 & 3.53 \\
    Pommes & 110 & 126 & 1.15 \\
    Burger & 105 & 122 & 1.16 \\
    Williamsbirne & 50 & 121 & 2.42 \\
    Alm-Breze & 100 & 114 & 1.14 \\
    Brotzeitkorb & 65 & 72 & 1.11 \\
    Kaesespaetzle & 92 & 100 & 1.09 \\
    Total & 1110 & 2696 & 2.43
\end{tabular}

\begin{figure}[ht]

\vskip 0.2in
\begin{center}
\centerline{
    \includegraphics[width=0.45\textwidth]{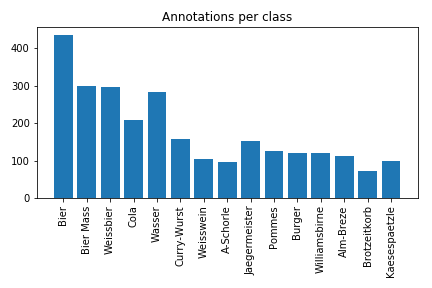}}
    \caption{Histogram showing distribution of the number of annotations per class.}
    \end{center}
\end{figure}
\begin{figure}[ht]
\vskip 0.2in
\begin{center}
    \centerline{
    \includegraphics[width=0.45\textwidth]{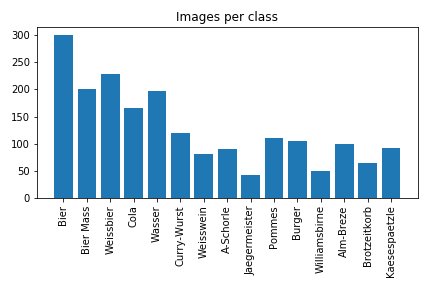}}
    \caption{Histogram showing distribution of the number of images containing certain class.}
    \end{center}
\end{figure}
\begin{figure}[ht]
\vskip 0.2in
\begin{center}
    \centerline{
    \includegraphics[width=0.45\textwidth]{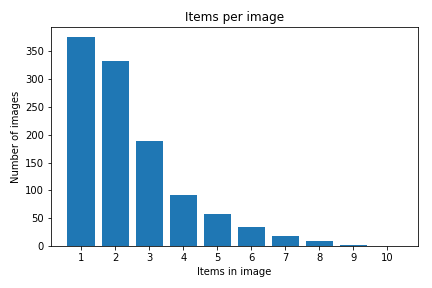}}
    \caption{Histogram showing distribution of the number of items per image.}
\end{center}
    \label{fig:class_dist}
\end{figure}
\begin{figure}[ht]
\vskip 0.2in
\begin{center}
\centerline{
    \includegraphics[width=0.5\textwidth]{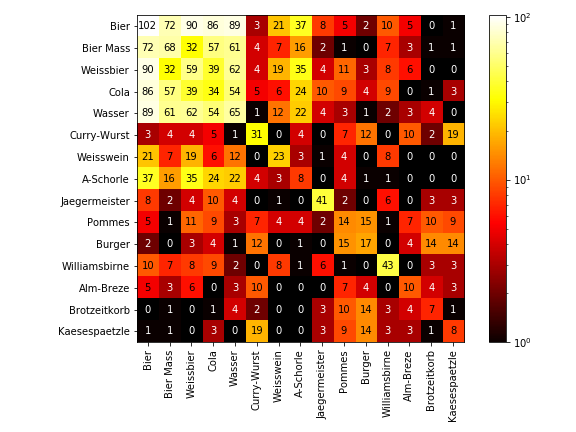}}
    \caption{Heat map of how often items appeared together on images.}
    \label{fig:heatmap}
\end{center}
\vskip -0.2in
\end{figure}

In addition, the dataset contains more than 6800 video frames labeled using object tracking.

\subsection{Dataset split}
We extracted 85 test images for evaluating the performance of the models. This test set also includes difficult cases like images with many objects which are close to each other, large occlusions by waiters and motion blur.
We also used the data from the last two days to evaluate the models trained on the data from the first 9 days, as well as other data splits.


\section{Evaluation metrics}
We used the standard evaluation metrics for object detection: precision and recall as well as the area under the precision-recall curve (AUC). Since our task was counting objects, we set the Intersection over Union (IoU) threshold to zero as it was not important to detect the location of the items but only their class.
\begin{figure}[ht]
    \centering
    \includegraphics[width=0.45\textwidth]{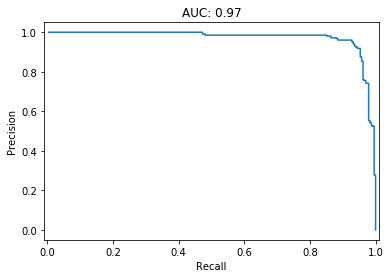}
    \caption{Example AUC curve of one of our models.}
    \label{fig:auc}
\end{figure}

\section{Benchmark}
We used the TensorFlow Object Detection API \cite{huang2017speed} for training the following object detection models on the image set: SSD, SSD+FPN (RetinaNet), RFCN. They were fine-tuned based on pre-trained models. We used the improved implementation of Faster RCNN \cite{chen2017spatial} for training the model.

In Table \ref{table:result_table} we list our results and measurements. \\
\begin{table}[ht]	
    \centering
    \begin{tabular}{l|lll}
        \thead{Approach} & \thead{Backbone} & \thead{AUC} & \thead{Speed [ms]} \\ \hline
        \makecell{SSD \\ \cite{liu2016ssd}} & Mobilenet & 0.86/- & 13\\
        \makecell{SSD + FPN \\ \cite{lin2017feature}} & Mobilenet & 0.98/0.92 & 29\\
        \makecell{RFCN \\ \cite{dai2016r}} & ResNet-101 & 0.97/0.89 & 64\\
	\makecell{Faster-RCNN \\ \cite{renNIPS15fasterrcnn}} & VGG-16 & 0.98/0.93 & 77
    \end{tabular}
    \caption{Model performance and speed. The first number in the AUC column denotes the result on the small test set, the second number - on the test set of data for the last two days (819 images for training, 222 images for testing).}
    \label{table:result_table}
   
\end{table}

\section{Further approaches}

\subsection{Self-Training}
In our dataset, only a small amount of available data is annotated since annotating all of the data is very labor-intensive. One way to increase the number of labels is to use self-training (a.k.a. pseudo labeling).

Pseudo labeling \cite{lee2013pseudo} uses a model trained on the labeled data to annotate the unlabeled data. On this automatically annotated data, a new model can be trained that would see more training data at the cost of possible errors in the annotations. 

\pagebreak

In our experiments, self-labeling did not improve results compared to models trained only on manually labeled data. Nevertheless, we provide code to create pseudo labels. 

\subsection{Limit training data}
In addition, we evaluated the performance of the chosen models for smaller training set sizes. Furthermore, we assessed the impact of training only on daytime images but testing on nighttime images having lower saturation. 

Results showed that performance was still good with only 60 samples per class. This suggests that despite us not having annotated much of the source data, it was already enough for this domain. In the second experiment (training on daytime, testing on nighttime images) performance dropped substantially, thus we need nighttime images in the training set for stable results. 

\subsection{Training on video frames}
Labels were automatically extended to neighboring frames using MedianFlow tracking. The average tracked sequence length was 22 frames. The annotations were manually reviewed, the frames for which the tracking failed were removed. We found it superior to firstly train RFCN on the original image set and then to fine-tune it on the video set for half of the iterations. The Faster RCNN implementation could be trained on the video frames directly. The increase in both the first-stage proposal recall (IoU 50\%) and the final AUC score was 3\% for both two-stage object detection models after also training them on the neighboring frames. Another implementation of RFCN based on \cite{chen17implementation} was used for the training.

\section{Conclusions}
Overall we show that deep learning is suitable for detecting objects at given environmental conditions. However, it is critical for automated checkout services that detection results are stable and on human performance level. In order to get closer to this goal, more research has to be conducted and/or more data needs to be annotated. The dataset we release gives a starting point. The source data with four camera angles could be exploited to achieve even better results in production.

\section*{Acknowledgements}
This work was done as a part of a student practical at Prof. G\"unnemann's chair at TUM. We would like to thank Daniel Z\"ugner for advising us and encouraging us to publish this dataset. Also credits to Peter Vogel who initiated this project and provided us with the data.

\bibliography{literature}
\bibliographystyle{icml2019}

\end{document}